%% file: acl_latex.tex
\pdfoutput=1

\documentclass[11pt]{article}

\usepackage[final]{acl}

\usepackage{times}
\usepackage{latexsym}
\usepackage{adjustbox}
\usepackage{arydshln}
\usepackage{xcolor}
\usepackage[T1]{fontenc}

\usepackage[utf8]{inputenc}

\usepackage{microtype}
\usepackage{multirow}
\usepackage[most]{tcolorbox}
\usepackage{inconsolata}

\usepackage{graphicx}
\usepackage{listings}
\usepackage{microtype}
\usepackage{inconsolata}
\usepackage{booktabs}
\usepackage{amsmath}
\usepackage{amssymb}
\usepackage{bbold}
\usepackage{multicol}
\usepackage{hyperref}

\newcommand{\eg}{\textit{e.g.}}
\newcommand{\ie}{\textit{i.e.}}

\title{BSharedRAG: Backbone Shared Retrieval-Augmented Generation \\for the E-commerce Domain}

\author{
Kaisi Guan
\quad Qian Cao  \quad Yuchong Sun \quad Xiting Wang \thanks{~~Corresponding authors.}  \quad Ruihua Song \footnotemark[1]
\\
Gaoling School of Artificial Intelligence, Renmin University of China, Beijing, China  \\
\texttt{\{guankaisi,caoqian4real,ycsun,xitingwang,rsong\}@ruc.edu.com}\\ 
}

\begin{document}
\maketitle

\input{tex/00-abs}
\input{tex/01-intro-sun}
\input{tex/02-related-sun}
\input{tex/05-method-sun}

\input{tex/06-experiment}

\input{tex/07-conclusion_and_appendix}

\end{document}

%% file: tex/00-abs.tex
\begin{abstract}

Retrieval Augmented Generation (RAG) system is important in domains such as e-commerce, which has many long-tail entities and frequently updated information. 
Most existing works adopt separate modules for retrieval and generation, which may be suboptimal since the retrieval task and the generation task cannot benefit from each other to improve performance.
In this paper, we construct a high-quality e-commerce dataset and use e-commerce as an example domain to show how to ensure effective transfer between the retrieval and generation tasks.
We propose a novel Backbone Shared RAG framework (\textbf{BSharedRAG}). It first uses a domain-specific corpus to continually pre-train a base model as a domain-specific backbone model and then trains two plug-and-play Low-Rank Adaptation (LoRA) modules based on the shared backbone to minimize retrieval and generation losses respectively. Experimental results indicate that our proposed BSharedRAG outperforms baseline models by 5\% and 13\% in Hit@3 upon two datasets in retrieval evaluation and by 23\% in terms of BLEU-3 in generation evaluation. 
Our codes, models, and dataset are available at \href{https://bsharedrag.github.io} {https://bsharedrag.github.io}.

\end{abstract}

%% file: tex/01-intro-sun.tex
\section{Introduction}
\label{sec:introduction}

Large Language Models (LLMs) have shown impressive performance in reasoning and remembering extensive knowledge \cite{gpt2,gpt3,openai2024gpt4,Claude,zhao2023survey,llama,llama2,yang2023baichuan,chen2024yulan}.
However, when facing domain-specific questions, especially the e-commerce domain that has many long-tail entities and frequently updated information, LLMs often have issues such as hallucinations~\cite{shi-etal-2023-hallucination,wang-etal-2023-hallucination,zhao-etal-2023-hallucination}.
Retrieval-Augmented Generation (RAG) has been proposed as an effective method for such cases~\cite{zhu2024rag_survey,gao2024ragsurvey2,zhao2024ragsurvey3}.
However, previous works \cite{zhang2023iag,shi2023replug,lin2023radit,zhang2024raft} mainly adopt a general-domain retriever and generator directly, which may be suboptimal due to the lack of domain-specific knowledge.
How to construct a domain-specific RAG system is an important research problem. To answer this question, we need to solve the following two challenges. 


\begin{figure*}[t]
\centering
\includegraphics[width=1\linewidth]{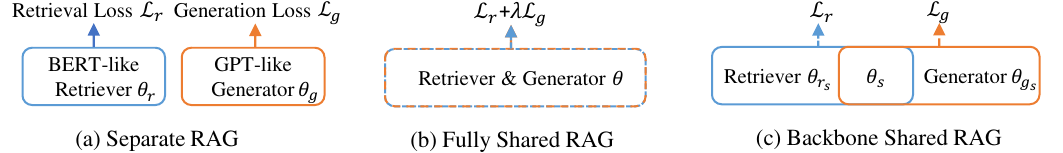}
\caption{Comparing three categories of possible RAG frameworks: (a) most previous works lie in the separate RAG category, in which the retrieval task and the generation task cannot benefit from each other; (b) only a few trials are about the fully shared RAG, which may suffer from performance decrease due to negative transfer and require effort-taking loss balancing to determine $\lambda$; (c) what we proposed is a backbone shared RAG, which ensures effective knowledge transfer between the two tasks without the need to perform effort-taking loss balancing.} 
\label{fig:archcomp}
\end{figure*}

\emph{C1 (dataset)}: How to build a high-quality RAG dataset with an informative knowledge base and correct question-answer (QA) pairs? For important domains like e-commerce, the existing knowledge bases are usually short and noisy, as most of them are from user-generated product reviews. The QA pairs are also of various qualities~\cite{deng-etal-2023-product} and may contain subjective or conflicting information because the answers are provided by users. 

\emph{C2 (framework)}: How to design a suitable RAG framework for domain adaptation? As shown in Figure~\ref{fig:archcomp} (a), existing works typically adopt separate modules for retrieval and generation \cite{shi2023replug,lin2023radit,zhang2023iag,langchain,Liu_LlamaIndex_2022}, for example, using BERT-like \cite{bert} embedding models for retrieving documents and an LLM for generating answers. 
In this design, retrieval and generation operate independently, hindering their potential to achieve optimal performance. It is often challenging for the retriever to leverage the continuous pre-training of the generator, while the generator's performance may be constrained by retrieved documents that, although similar to the query, are less effective in addressing the questions (Section~\ref{sec:retrieval:analysis}).
Although we can build the retriever and generator both on a fully shared LLM and optimize it with multi-task learning as shown in Figure~\ref{fig:archcomp} (b), it can suffer from negative transfer between tasks, \ie, performance decrease due to the potential conflicts in the retrieval task and the generation task. Moreover, balancing the retrieval loss and the generation loss is non-trivial and may lead to an effort-taking hyperparameter search.



To address the two challenges, we first propose the WorthBuying dataset with 735K high-quality documents, 50K Question-Document-Answer (QDA) tuples, and human annotated test data of relevant documents for 1K questions and 500 QA pairs (\emph{C1}).
The knowledge base in our dataset comes from professional users, reducing conflicts and errors, and is more informative, with 1.1K words per document rather than a few dozen words as in existing e-commerce knowledge bases. We also annotate high-quality QA pairs with GPT-4~\cite{openai2024gpt4} and manually review the test set.
We then propose a specifically designed BSharedRAG framework that effectively adapts RAG to the e-commerce domain (\emph{C2}).
As shown in Figure~\ref{fig:archcomp} (c) we apply two task-specific modules to independently minimize the retrieval loss and generation loss, which avoids effort-taking loss balancing.
With the shared backbone that benefits from domain-specific continual pre-training, the retriever and the generator can benefit from each other to improve both retrieval and generation performance.\looseness=-1


Our contributions are summarized as follows:

    $\bullet$ We construct a high-quality dataset called \textbf{WorthBuying} for the e-commerce domain, which contains 735K documents, 50K QDA tuples for training and some high-quality human labeled test data to facilitate further research.
    
    $\bullet$ We propose a backbone-shared RAG framework (\textbf{BSharedRAG}) for the e-commerce domain, which enables effective knowledge transfer between the retrieval task and the generation task to improve the performance of both tasks.
    Although we focus on e-commerce, the framework can be generalized to other domain-specific RAG models. 
    
    $\bullet$ Experimental results demonstrate that our BSharedRAG, with affordable efficiency, outperforms traditional RAG methods in both retrieval and generation evaluation.

%% file: tex/02-related-sun.tex
\section{Related Work}

\paragraph{LLMs for the E-Commerce Domain.}
To adapt LLMs for the e-commerce domain, some studies leverage LLMs like GPT-3.5-turbo and GPT-4 to generate e-commerce instruction tuning datasets~\cite{li2024ecomgpt,peng2024ecellm}. 
Continual pre-training on e-commerce data has been shown to further enhance LLM performance in e-commerce tasks~\cite{ma2023ecomgptct}. Despite these advancements, issues such as hallucination hindering real-world application, especially for Product Question Answering (PQA) tasks that heavily rely on product knowledge~\cite{miller2020reviewqa,shi2023llamae,li2024ecomgpt}. Existing PQA datasets only contain short, non-informative knowledge bases about products, and QA pairs are of various quality~\cite{amazonqa,JD,shen2023xpqa,deng-etal-2023-product,Taobao}, since they contain subjective or conflicting answers provided by users.
In this paper, we address the aforementioned issues by proposing a high-quality WorthBuying dataset. Moreover, we propose a BSharedRAG framework to improve retrieval and generation performance significantly.

\paragraph{Retrieval Augmented Generation for LLMs.}
Retrieval Augmented Generation (RAG), where the retriever provides relevant knowledge from external sources and LLMs answer the query based on the retrieved results, significantly reduces problems such as hallucinations and untimely knowledge updating of LLMs~\cite{zhu2024rag_survey,zhao2024ragsurvey3,gao2024ragsurvey2}.  
\begin{figure*}[t]
\centering
\includegraphics[width=\linewidth]{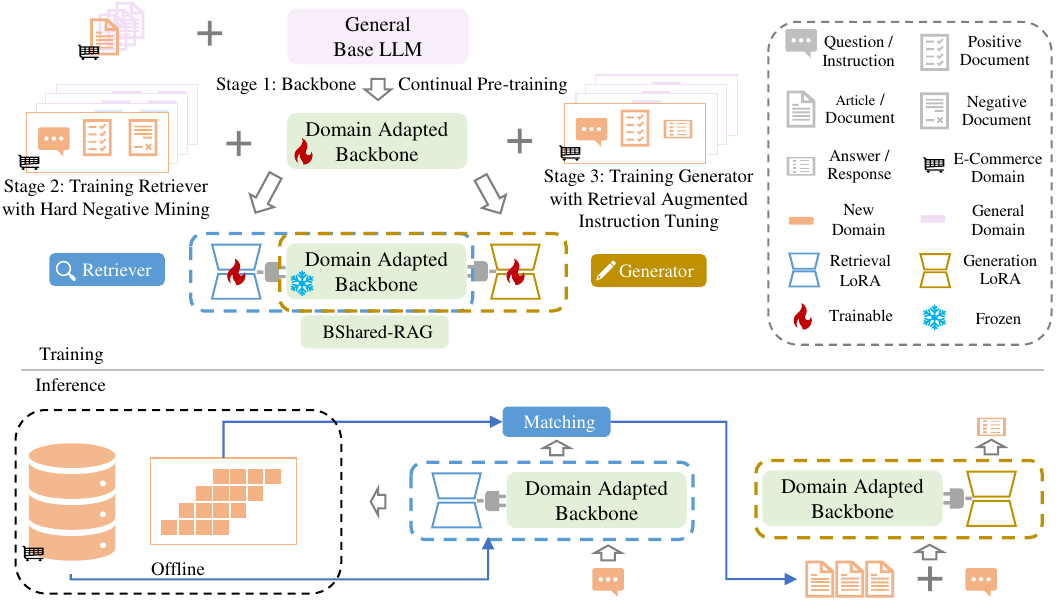}
\caption{Overview of training and inference of our proposed BSharedRAG Framework. 
}
\label{fig:framework}
\end{figure*}
In previous SeperateRAG frameworks, the retriever and the generator are typically separate and share no parameters, with the retrievers being off-the-shelf embedding models like DPR~\cite{dpr}, BGE~\cite{bge}, E5~\cite{E5}, or BERT-like architectures~\cite{bert}, and the generators being LLMs~\cite{llama,llama2,yang2023baichuan}. Some RAG frameworks, such as RETRO \cite{borgeaud2022retro}, REPLUG \cite{shi2023replug}, and RA-DIT \cite{lin2023radit}, make some performance improvements within the SeperateRAG framework. These frameworks require an independent retriever alongside the language model, without any shared parameters between the retriever and generator, creating a gap between these components. It is challenging for them to share the benefits derived from continual pre-training (CPT), making it difficult for them hard to adapt to a specific domain.
Recently, GritLM~\cite{muennighoff2024gritlm} is proposed, in which the retriever and generator share all parameters and thus the the retriever can benefit from the capacity of LLM. However, it is unclear how GritLM can be adapted to domain-specific scenarios: it requires large-scale training data and high computation costs that can hardly be satisfied in a specific domain. In comparison, we propose a lightweight framework that can efficiently and effectively convert a continually pre-trained domain-specific LLM to a backbone-shared domain-specific RAG model. 


%% file: tex/05-method-sun.tex
\section{Method}
In this section, we introduce our BSharedRAG for scenarios that require extensive domain knowledge, such as e-commerce. 

\subsection{Framework}
\textbf{Separate RAG}. Traditional RAG frameworks treat retrieval and generation as two distinct tasks, where the retriever and generator share no parameters (Figure~\ref{fig:archcomp}(a)).
Formally, they optimize two groups of parameters separately:
\begin{equation}
    \min\ \mathcal{L}_r(\theta_{r} ) , \ \   \min \mathcal{L}_g(\theta_{g})
\end{equation}
where $\theta_{r}$ and $\mathcal{L}_r$ are the parameters and loss for retrieval model, and $\theta_{g}$ and $L_g$ are those for generation model.
This paradigm cannot benefit from the shared information between the two related tasks as discussed in Section~\ref{sec:introduction}, which limits the performance of both retrieval and generation. 

\noindent \textbf{Fully Shared RAG}.
One straightforward method to overcome the problems is to build a shared model that can be used for retrieval and generation based on multi-task learning (Figure~\ref{fig:archcomp} (b)):
\begin{equation}
    \min \mathcal{L}_r(\theta) + \lambda \mathcal{L}_g(\theta)
\end{equation}
where $\theta$ is the parameters of the shared model, and $\lambda>0$ is the weight for balancing the two losses. 
This framework may suffer from negative transfer, i.e., performance decrease in both tasks due to the potential conflicts between them, as verified in our experiments. Moreover, finding an appropriate $\lambda$ to ensure loss balancing is non-trivial and may require an effort-taking hyperparameter search.

\noindent \textbf{Backbone Shared RAG}.
To make both retrieval and generation benefit from the capability of LLM, and avoid the difficulty of balancing two objectives, we propose a BSharedRAG framework.
The basic idea is to use a frozen LLM backbone as a shared information processor and two task-specific modules for retrieval and generation tasks.We use LoRA \cite{hu2021lora} method to finish this idea.
Under this design, the two tasks can share knowledge from the backbone, \eg, domain-specific continual pre-training, while using task-specific parameters for different purposes to avoid the difficulty of balancing two tasks and negative transfer:

\begin{equation}
\label{eq:optgoal}
    \min \mathcal{L}_r(\theta_{r_s}; \theta_s), \ \    
    \min \mathcal{L}_g(\theta_{g_s}; \theta_s)
\end{equation}
where $\theta_{r_s}$ and $\theta_{r_s}$ are task-specific modules unpon a pre-trained shared backbone model $\theta_s$.


\subsection{Training Strategies}
As shown in Figure~\ref{fig:framework},
we adopt the following three training stages to optimize Equation~(\ref{eq:optgoal}). 


\paragraph{Stage 1: Backbone Continual Pre-training.}
We leverage continual pre-training to adapt a general domain LLM (Baichuan2-7B-Base \cite{yang2023baichuan} here) to a specific domain such as e-commerce (here called Ecom-base). 
The domain-specific continual pre-training objective is the next token prediction task:
\begin{equation}
    \max\limits_{\theta_s}\ \sum_{i=1}^N \mathrm{log}\ p(y_i|y_{<i};\theta_s)
\end{equation}
For the training data, we mix corpus from the e-commerce domain and the general domain to avoid catastrophic forgetting following~\cite{xie2023efficient,ke2023continual,que2024dcpt}.

\input{tables/dataset}

\paragraph{Stage 2: Training Retriever with Hard Negative Mining.}

We then minimize $\mathcal{L}_r$ by fine-tuning a domain-specific retrieval model based on a frozen base LLM $\theta_s$ with trainable LoRA \cite{hu2021lora} module $\theta_{r_s}$. 
As our retrieval model adopts a GPT-like architecture, we adhere to the RepLLaMA paradigm \cite{ma2023repllama} to derive sentence embeddings from the End of Sentence (EOS) token.

Following the approach of the bi-encoder dense retriever DPR \cite{dpr}, we employ in-batch contrastive learning to fine-tune Ecom-base with InfoNCE loss:

\begin{equation}
\mathcal{L}_r = -\log\ \frac{e^{sim(q,D^+)}}{e^{sim(q,D^+)}+\sum\limits_{D_i^- \in {D^-}}e^{sim(q,D_i^-)}}
\end{equation}

Here, $D^+$ represents the relevant document, and ${D^-}$ denotes a set of irrelevant documents to the query. We employ the in-batch negatives method, where one positive document and several negative documents are included in one sample.

We also propose a strategy for identifying hard negatives during contrastive learning. Specifically, we utilize a baseline retrieval model to retrieve a preliminary set of potential negatives, then the high-ranking yet erroneously matched documents, \ie, {hard negatives}, which helps refine the model’s discrimination capabilities.

For our hard-negative mining, we first employ a base retrieval model that is fine-tuned on random negative samples to rank documents. Then we utilize the top 30 but incorrect results as hard-negative samples. Experimental results show that this strategy enhances the effectiveness of retrieval.LoRA is applied to all layers of our LLMs. The targeted weights for LoRA adaptation consist of the query, key, value, and output weights $(W_q, W_k, W_v, W_o)$ within the attention modules, as well as the weights in the multi-layer perceptron (MLP) components. 

\paragraph{Stage 3: Training Generator with Retrieval Augmented Instruction Tuning.}
To ensure that the generator can effectively utilize retrieved information, we augment the instruction-response pairs by incorporating the retrieved articles, forming triples $(q, d, a)$, where $q$ denotes question tokens, $d$ denotes the retrieved document, and $a$ denotes the answer. 
Specifically, 
\begin{equation}
  \mathcal{L}_g =   
  - \sum_{i=1}^N \mathrm{log}\ p(a_i|a_{<i},q,d,\theta)
\end{equation}
Similar to the embedding model, we employ the LoRA module to fine-tune only a subset of additional parameters $\theta_{g_s}$ on top of foundational LLM.

\subsection{Inference}
During inference, the retriever is first employed to obtain documents, which are subsequently given to the generator for processing (Figure~\ref{fig:framework}).
Specifically, we first prepare all embeddings for the documents in the domain-specific knowledge base by using the BSharedRAG retriever offline. In the online stage, we only extract the embedding of the question and use it to retrieve the most relevant $N$ documents ($N=3$). Finally, we use the BSharedRAG generator to obtain an answer to the question by taking the top documents as input.

\section{WorthBuying Dataset}

Existing e-commerce datasets are mainly constructed from user-generated product reviews, which are brief and noisy, resulting in limited assistance to LLMs in generating rich content~\cite{icdm16-amazon,amazonqa}.
To solve this issue, we build a WorthyBuying dataset, which includes an accurate knowledge base and high-quality QDA tuples for the e-commerce domain.

\paragraph{Data Collection.}
To obtain a high-quality corpus for the e-commerce domain, 
we collect
1.1M professional product reviews from an e-commerce website\footnote{\href{https://www.smzdm.com/}{https://www.smzdm.com/}}.
Beginning from the raw corpus, we conduct several steps of data cleaning to ensure the quality, such as filtering contents with dirty words\footnote{\href{https://github.com/LDNOOBW/List-of-Dirty-Naughty-Obscene-and-Otherwise-Bad-Words}{https://github.com/LDNOOBW/List-of-Dirty-Naughty-Obscene-and-Otherwise-Bad-Words}.}, removing any HTML tags and images, and deleting those with the length beneath a threshold.
We then adopt the Minihash method\footnote{\href{https://github.com/ChenghaoMou/text-dedup}{https://github.com/ChenghaoMou/text-dedup}} to eliminate duplicate entries, which finally results in 735k cleaned documents.
More details are presented in the Appendix.

\paragraph{QDA Tuples Generation.}
Users often ask LLMs to obtain information about products in the e-commerce domain. 
However, the questions and answers in existing datasets are often derived from brief user QA comments, which are neither verified nor grounded in any relevant documents.
The informative review documents we collected can provide sufficient clues to answer product-related questions.
To facilitate documents lacking high-quality QDA tuples aligned with user needs, we leverage GPT-4 to generate questions and answers by prompt with these documents. We collect 50K such $\langle\text{question}, \text{document}, \text{answer}\rangle$ tuples grounded on documents.
This dataset can support the training of the embedding model and generation model.

\paragraph{Dataset Statistics and Comparisons.}
As indicated in Table~\ref{tab:dataset}, previous datasets consist of undetailed short documents.
Their questions and answers are directly collected from brief user QA comments without verification, leading to noise and even incorrect answers.
Our WorthBuying dataset contains much longer documents with more than 50k QDA tuples grounded on 735k informative documents.
We also contain the most detailed product categories than existing datasets.

%% file: tables/dataset.tex
\begin{table*}
\setlength{\abovecaptionskip}{5pt}   
\setlength{\belowcaptionskip}{5pt}
    \centering
    \scalebox{0.85}{
    \begin{tabular}{ccccccc}
    \toprule
      Dataset  &  Document Source & \# Categories & \# Document & Avg words per Doc &  Release\\
      \midrule
      AmazonQA \cite{ijcai19-amazonqa}&PR & 17 & 923K &63.41 & $\checkmark$\\
      SubjQA \cite{emnlp20-subjqa} &PR  & 6 & 10,098 &289.87  &  $\checkmark$\\
      semiPQA \cite{ecnlp22-semipqa}  & PI  & - & 11,243 & -& $\times$\\
      xPQA \cite{shen-etal-2023-xpqa} &PR  & - & 2,500 & 76.87 & $\checkmark$\\
      JD \cite{wsdm19-answer-gen-gao} &PR \& PI & 38 & 469,955  &16.94 & $\checkmark$\\
      Taobao \cite{wsdm19-answer-gen-chen} &PR & 2 & 1,155,530 &74.15 & $\times$\\ \midrule 
WorthyBuying (ours) & PA & 25 \& 128& 735,937  & \textbf{1171.25} &  $\checkmark$ \\
      \bottomrule
    \end{tabular}}
    \caption{Comparison of Product Question Answering (PQA) datasets. 
    The document types are classified into Product Reviews (PR), Product Information (PI), and \textbf{Product Analysis from professional users (PA) }. 
    }
    \label{tab:dataset}
\vspace{-0.3cm}
\end{table*}

%% file: tex/06-experiment.tex
\section{Experiment}

\input{tables/retrieve}
\subsection{Experiment Setup}
\subsubsection{Datasets and Evaluation Metrics}

To evaluate the retrieval effectiveness, we use an existing dataset \textbf{CPR-Ecom}~\cite{long2022multicpr}, which contains a corpus of 1,002,822 Chinese documents, 100k human annotated query and relevant document pairs for training and 1k for testing in an e-commerce domain, and our constructed \textbf{WorthBuying} dataset, which contains a corpus of 735k Chinese documents, 50k GPT-4 generated questions from 50k documents for training and 1k GPT-4 generated questions with 10.2k human-annotated documents with 1 or 2 relevance ratings for testing, as benchmarks. We adopt two retrieval metrics for evaluating our models: nDCG (Normalized Discounted Cumulative Gain) and Hit Rate.
nDCG@n measures the ranking quality of top n retrieved documents, considering accumulated gains from relevant documents while discounting their positions. Hit@n assesses the presence of relevant documents within the top n results.

To evaluate generation quality, we use our built WorthBuying dataset. We randomly sample 500 QA-pairs from the test set and hire human annotators to verify and correct the questions and answers.
This manual correction ensures the accuracy of the results and provides a more reliable evaluation. 
We employ a comprehensive set of widely used automatic metrics for evaluation: n-gram based BLEU-3~\cite{bleu} and ROUGE-L~\cite{rouge}, BERTScore~\cite{bertscore} which calculates semantic similarity of ground-truth and generated answers based on a BERT model.
Accuracy~\cite{es2023ragas} is calculated by GPT-4 to evaluate question and answer pairs.
More details are in the Appendix~\ref{sec:expr_details}.

\input{tables/gen}

\subsubsection{Baselines}
We compare our BSharedRAG Retriever with several widely-used baseline retrieval models:

\noindent \textbf{BGE}~\cite{bge} is the state-of-the-art BERT-like model trained with multi-stage contrastive learning using over 200M paired data.
To ensure a fair comparison, we fine-tune BGE through continual pre-training and in-domain contrastive learning with hard negative samples, resulting in two variants.

\noindent \textbf{M3E}~\cite{m3e} is an open-source text embedding model trained by a massive corpus with over 22M sentence pairs, supporting bilingual (Chinese-English) text retrieval.

\noindent \textbf{Multilingual-E5}~\cite{wang2024multie5} transfers E5 embedding model (English only)~\cite{E5} to 100+ languages with a two-stage training and exhibits competitive performance across a broad range of languages.


For generation, we choose the following representative LLMs from both open domains and e-commerce domain as baselines for comparison: 


\noindent \textbf{GPT-3.5} and \textbf{GPT-4} \cite{openai2024gpt4} are developed by OpenAI which are the most advanced LLMs. We feed them with retrieved passages in the prompt for a fair comparison.

\noindent \textbf{EcomGPT}~\cite{li2024ecomgpt} is an e-commmerce domain-specific LLM trained on BLOOMZ-7B~\cite{muennighoff2022bloomz} with their close-source instruction dataset EcomInstruct.

\noindent \textbf{Baichuan2-7B-Chat}~\cite{yang2023baichuan} is one of the best open-source Chinese LLMs. It adopts 100k samples for supervised fine-tuning, after which is optimized using Reinforcement Learning from Human Feedback (RLHF)~\cite{ziegler2020rlhf}.


\noindent \textbf{Seperate RAG} has separate parts for retrieval and generation tasks, as shown in \ref{fig:archcomp}(a). To make sure the same parameters as BSharedRAG in two stages, we train Llama3-8b \cite{llama3}, a powerful foundation LLM and eCeLLM-M(7B) \cite{peng2024ecellm}, a powerful e-commerce domain LLM with the same data, and configuration as BSharedRAG-retriever in retrieval stage. For the generation, we use the BSharedRAG-generator.

\noindent \textbf{Fully Shared RAG} integrates retrieval and generation into a single model, as shown \ref{fig:archcomp}(b). 
Using the same base model, data, and configuration as BSharedRAG, it employs a single LoRA to fully share retrieval and generation parameters.

\subsection{Experimental Results}

\subsubsection{Retrieval Evaluation}
Results in Table~\ref{table:retrieve} indicate that our \textit{BShared-RAG Retriever} significantly outperforms all baselines in all metrics upon both datasets. 
\textit{BGE-large-zh+HN}, which is fine-tuned on our dataset using hard negative contrastive learning, serves as the strongest baseline. However, our model still achieves improvements ranging from 5\% to 17\% over this baseline.
The \textit{FullyShared-RAG} performs much worse than ours and most baselines in separate RAGs. This indicates that it is difficult to optimize two objectives with fully shared parameters. Compared to BERT-like retrievers, \eg, BGE-large-zh, our proposed method can effectively benefit from continual pre-training. In contrast, \textit{BGE-large-zh+CPT+HN} has significant drops in all metrics when applying continual pre-training. 
In our analysis, there is evidence indicating that embedding models based on architectures like BGE require more fine-tuning training data after CPT, potentially reaching the terabyte (TB) level \cite{wang2024e5_instruct}. Fine-tuning with the same amount of data as BSharedRAG is insufficient for the BGE model, which could lead to a decline in performance.
We conduct ablation studies to analyze the detailed contributions of each module. 
Both CPT and HN contribute to significant and consistent improvements, while CPT brings more gains.

\subsubsection{Generation Evaluation}
\begin{figure}
    \centering
    \includegraphics[width=0.99\linewidth]{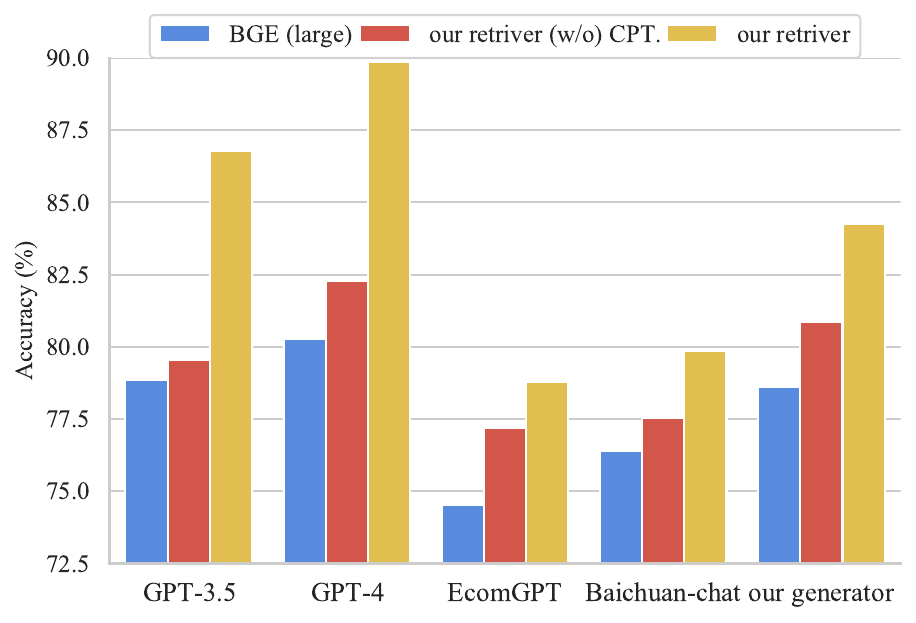}
    \caption{Evaluating the influence of different retrievers to generation effectiveness. CPT is continual pre-training, which benefits retrieval effectiveness a lot via sharing an LLM backbone. Accuracy is judged by GPT-4. Other metrics are not shown due to limited space, but we observe similar trends.}
    \label{fig:accuracy}
    \vspace{-3mm}
\end{figure}
Results in Table~\ref{table:gen} show that our model performs the best among open-source models. In terms of ROUGE-L and BertScore, our model is even slightly better than GPT-3.5. 
FullyShared-RAG performs the worst because it cannot well balance the retrieval and generation tasks.
Baichuan2-7b-base has no instruction following abilities and thus it performs lowest in QA testing. After RAG-IT, it can obtain instruction following abilities and thus its performance is significantly improved, \eg, from 76.81\% to 82.91\% in Accuracy. A strong chat model, such as Baichuan2-7b-chat, can be also improved by RAG-IT because it brings domain-specific knowledge.
Our proposed BShared-RAG generator performs better than \textit{Baichuan-7b-base + RAG-IT} by 37\% in BLEU-3 and 31\% in ROUGE-L. 
These results demonstrate CPT and RAG-IT are both necessary and effective to improve generators.
Meanwhile, the Separate-RAG model, which employs Llama3 and EcellM as retrievers, maintains the same parameters as BSharedRAG but exhibits lower performance. This result underscores the importance of sharing a unified backbone, confirming its necessity for enhanced performance.

\begin{figure*}[t]
\centering
\includegraphics[width=1\linewidth]{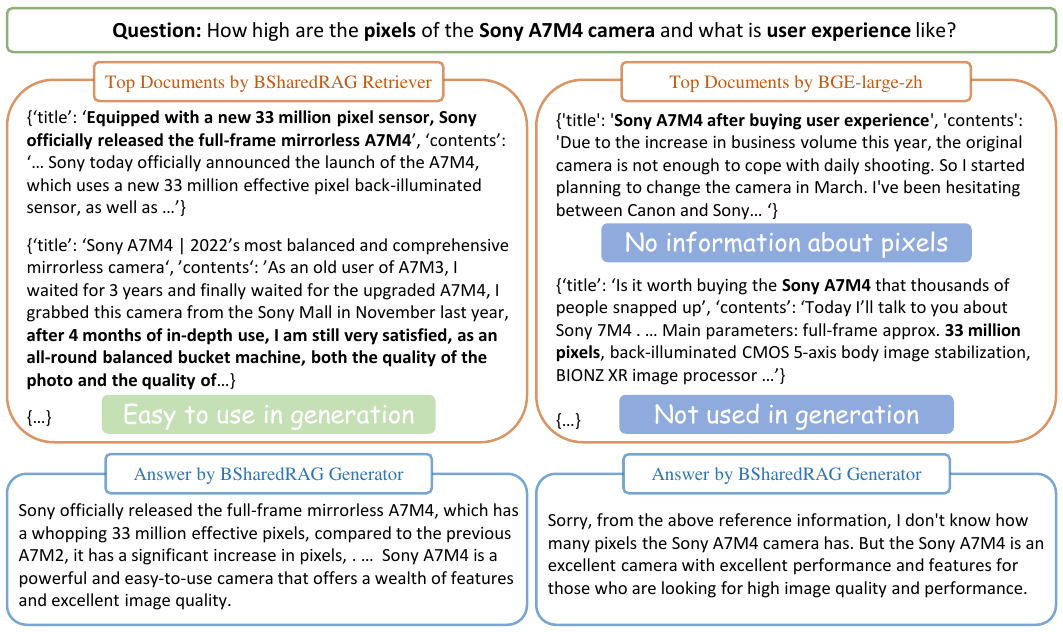}
\caption{A representative example to compare our BSharedRAG with a separate RAG. For the given question, our BSharedRAG Retriever favors the documents, in which some sentences are easy to be generated from the prompt of question. In contrast, the BERT-like BGE-large-zh model tends to retrieve some documents, in which some sentences match the question well. However, such document may be less suitable for generating answers due to some issues, \eg, important information missing or not easy to be used by generators. }
\label{fig:casestudy}
\end{figure*}

\input{tables/correlation}

\subsubsection{Influence of Retrieval to Generation}
\label{sec:retrieval:analysis}
We further conduct some experiments to analyze how retrievers influence generation. We compare the accuracy of generated results by models using BGE-large-zh, our BShared-RAG retriever and BShared-RAG w/o CPT. Results in Figure~\ref{fig:accuracy} indicate that our BShared-RAG Retriever can significantly improve generation accuracy no matter using GPT-3.5/GPT-4 or open-source models. Without CPT, the performance has a large drop. This underscores how important it is to bring continual pre-training into retrieval models. 

\subsection{Analysis of BSharedRAG Performance Gain}
Reasons for BSharedRAG’s outstanding performance can be attributed to following three factors:

\textbf{BSharedRAG provides a way to scale up the retriever parameters appropriately}. Conventional BERT-based embedding models \cite{bert,E5,bge} face inherent architectural limitations that hinder their ability to scale up parameters, thus preventing them from fully leveraging the benefits of scaling laws. While some approaches have employed LLM as retrievers, these methods often consume excessive memory. BSharedRAG effectively addresses these challenges by sharing a common backbone between the retriever and generator, allowing the retriever to benefit from larger parameter sizes without increasing the overall computational overhead.

\textbf{BSharedRAG enables both the retriever and generator to benefit from continued pre-training}. In SeparateRAG, the retriever and generator are trained independently, meaning the benefits of continued pre-training cannot be shared between them. On the other hand, in the FullySharedRAG structure, the retrieval loss and generation loss can interfere with each other, leading to degraded performance. In contrast, the BSharedRAG framework requires only continued pre-training (CPT) on the backbone LLM, allowing both the retriever LoRA and generator LoRA to benefit from domain-specific performance improvements achieved through CPT.

\textbf{BSharedRAG facilitates better preference alignment between the retriever and generator}. To verify this, we conducted an experiment. For each question, we gathered the ground-truth document title and top non-ground-truth titles returned by two methods. We then concatenated each question with a document title and calculated the generation probability using our BSharedRAG Generator. Next, we computed Kendall’s tau and Pearson correlation between ranked lists from either BGE-large-zh or our retriever, and from the generator. As shown in Table~\ref{table:correlation}, our method achieves significantly higher correlations, indicating that the BSharedRAG Retriever favors texts with higher generation probabilities, helping it outperform separate RAG baselines.



As illustrated in Figure~\ref{fig:casestudy}, BGE-large-zh prioritizes titles closely matching the question, whereas our BShared-RAG Retriever ranks titles that better answer the question.
For example, our method ranks ``Equipped with a new 33 million pixel sensor, Sony officially released the full-frame mirrorless A7M4'' higher, which is like a more relevant answer to the question.
In contrast, BGE-large-zh ranks ``Sony A7M4 after buying user experience'' without ``pixel'' information higher, or another document the second, which is hard for the generator to capture due to its isolation in the middle~\cite{liu2024lost}.

\input{tables/space}

\subsection{Inference space cost and time}
We evaluate the inference space cost and time of our BSharedRAG framework. 
As demonstrated in Table \ref{table:space}, BSharedRAG has a lower space cost than SeperateRAG due to eliminating the need for an additional retrieval component. 
In the retrieval part, BSharedRAG integrates a shared backbone model, which results in a marginal increase of retriever parameters compared to the BERT-based model. However, this does not lead to a significantly longer inference time. As shown in Table \ref{table:time}, the overall time increase is just 0.34 seconds, which remains within an acceptable range.
\input{tables/time}

%% file: tables/retrieve.tex
\begin{table*}[ht]
\centering
\scalebox{0.78}{
\begin{tabular}{lcccccccc}

\toprule
\multirow{2}*{\textbf{Model}} & \multicolumn{4}{c}{CPR-Ecom} &  \multicolumn{4}{c}{WorthBuying} 
\\\cmidrule(r){2-5} \cmidrule(r){6-9} 
~ & nDCG@3 & Hit@3 & nDCG@5 &Hit@5 & nDCG@3 & Hit@3 & nDCG@5 &Hit@5 \\ 
\midrule 
Multi-E5-large \cite{wang2024multie5} & 42.44& 49.20 & 45.36& 56.30&34.98 &39.50 & 37.51 & 45.20\\
M3E-large \cite{m3e} &44.89& 51.70 &47.65 &58.40 &43.02 &49.40 & 45.64 & 55.40\\
BGE-large-zh \cite{bge} & 51.25 & 58.60 & 54.44 & 66.30 & 47.01 & 53.50 & 49.86 & 56.30 \\
BGE-large-zh + HN & \underline{52.38} & \underline{60.35} & \underline{56.23} & \underline{68.50} & \underline{48.46} & \underline{55.70} & \underline{50.83} & \underline{58.30} \\
BGE-large-zh + CPT + HN & 47.15 & 58.10 & 50.28 & 61.20 & 45.38 & 51.20 & 46.36 & 55.40 \\
FullyShared-RAG & 38.67 & 44.80 & 43.24 & 48.80 & 33.28 & 36.70 & 36.56 & 41.20 \\
\cmidrule(lr){1-9}
\multirow{2}*{\textbf{BShared-RAG Retriever (ours)}} & \textbf{61.39} & \textbf{68.40}&\textbf{64.28}&\textbf{75.40} & \textbf{51.83}& \textbf{58.60}&\textbf{54.06} & \textbf{63.90}\\
~ & (+17\%) & (+13\%) & (+14\%) & (+10\%) & (+7\%) & (+5\%) & (+6\%) & (+10\%)\\
\cmidrule(lr){1-9}
~~~ w/o CPT & 56.74 & 63.80 & 58.80 & 67.90 & 49.02 & 54.50 & 51.78 & 58.30 \\
~~~ w/o HN & 58.35 & 64.70 & 61.23 & 70.60 & 49.34 & 56.30 & 52.03 & 60.90 \\ 
~~~ w/o CPT \& HN & 53.84 & 61.50&	57.76&	66.30&	47.24&	53.40&	49.93&	57.20 \\ 
\bottomrule
\end{tabular}}

\caption{Comparing retrievers of different RAG frameworks. CPT denotes continual pre-training and HN denotes using hard negative samples. 
Our BShared-RAG Retriever outperforms all baselines by a large margin. CPT fails to help the BGE adapt to the e-commerce domain and even hurts the performance. FullShared-RAG performs the worst, showing that sharing all parameters between retrieval and generation leads to severe performance degradation.
}
\label{table:retrieve}
\end{table*}

%% file: tables/gen.tex
\begin{table*}[ht]
\centering
\small
\begin{tabular}{lccccc}

\toprule
 \textbf{Model} & \textbf{BLEU-3} & \textbf{ROUGE-L} & \textbf{BERTScore} & \textbf{Accuracy (\%)} \\
\midrule 
\textit{FullySharedRAG as retriever} \\
\text{FullySharedRAG} & 2.35 & 2.58 & 62.37  & 68.34 \\
\cmidrule(lr){1-5}
\textit{SeperateRAG as retriever} \\
\text{SeperateRAG (llama3 retriever)} & 10.35 & 10.38 & 71.84  & 81.38 \\
\text{SeperateRAG (ecellm retriever)} & 10.13 & 11.23 & 71.98  & 82.41 \\
\cmidrule(lr){1-5}

\textit{BSharedRAG Retriever as retriever}\\
GPT-3.5 & \textcolor{gray}{16.07} & \textcolor{gray}{12.76} & \textcolor{gray}{74.33} & \textcolor{gray}{86.77} \\
GPT-4~\cite{openai2024gpt4} & \textcolor{gray}{19.76} & \textcolor{gray}{15.92} & \textcolor{gray}{77.98} & \textcolor{gray}{89.86}\\
EcomGPT~\cite{li2024ecomgpt} & 6.91 & 8.12& 69.18 & 78.70 \\
Baichuan2-7b-base & 2.32 & 2.49 & 62.56& 76.81\\
Baichuan2-7b-base + CPT & 2.46 &3.24 & 62.58 & 80.85\\
Baichuan2-7b-base + RAG-IT &9.24 & 9.83 & 70.07 & \underline{82.91} \\
Baichuan2-7b-chat~\cite{yang2023baichuan} & 8.82 & 11.21 & \underline{72.45} & 79.84 \\
Baichuan2-7b-chat + RAG-IT & \underline{10.24} & \underline{12.34} & 72.12  & 82.33 \\
\textbf{BSharedRAG Generator (ours)}  & \textbf{12.63 (+23\%)} &\textbf{12.85 (+4\%)} &\textbf{74.75 (+3\%)} & \textbf{84.26 (+2\%)} \\

\bottomrule
\end{tabular}

\caption{Evaluation of generation results based on different retrievers on the WorthBuying-PQA test set. RAG-IT denotes retrieval augmented instruction tuning. The FullySharedRAG method performs worse because the generation objective may conflict with the retrieval objective. Compared with Baichuan2-7b series of baselines, our model achieves the best performance, demonstrating both CPT and RAG-IT contribute to the final performance.}
\label{table:gen}
\end{table*}

%% file: tables/correlation.tex
\begin{table}[!t]
\centering
\renewcommand{\arraystretch}{1.5}
\setlength\tabcolsep{5pt}
\small
\scalebox{0.95}{
\begin{tabular}{ lcccc } 
\toprule
\textbf{Retrieval Model} & \textbf{Kendall's $\tau$} & \textbf{Correlation} \\ \cmidrule(lr){1-5}
BGE-large-zh & 0.0498 & 0.045  \\
Llama3-retriever&	0.0937&	0.085 \\
Ecellm-retriever&	0.1032&	0.119 \\
BSharedRAG Retriever & 0.1433 & 0.155 \\ 
\bottomrule
\end{tabular}}
\caption{Measuring whether a retriever favors a sentence, which is easy to generate from a prompt of a question. We calculate Kendall's tau and Pearson's correlation between two ranked lists, one by a retrieval model and the other by the generation probability estimated by BSharedRAG Generator. The larger the better.} 
\vspace{-2mm}
\label{table:correlation}
\end{table}

%% file: tables/space.tex
\begin{table}[!t]
\centering
\renewcommand{\arraystretch}{1.5}
\setlength\tabcolsep{2pt}
\small
\scalebox{0.77}{
\begin{tabular}{ lcccc } 
\hline
\textbf{Model} & \textbf{Parameters} & \textbf{RAM}& \textbf{VRAM} & \textbf{Storage}\\ \hline
SeperateRAG (bge-zh-large) & 7.87B & 30.07GB & 58.63GB & 15.19GB \\
SeperateRAG (llama3-retriever) & 15.84B & 61.19GB & 120.38GB & 29.05GB \\
\textbf{BSharedRAG} & \textbf{7.54B} & \textbf{29.02GB} & \textbf{56.61GB} & \textbf{14.09 GB} \\ 
\bottomrule
\end{tabular}
}
\caption{Comparing the space costs of different RAG frameworks, SeperateRAG (bge-zh-large) and SeperateRAG (llama3-retriever) respectively utilize the bge-zh-large and finetuned llama3 models,  as retrievers, while sharing the same generator as BSharedRAG.
}
\vspace{-2mm}
\label{table:space}
\end{table}

%% file: tables/time.tex
\begin{table}[!t]
\centering
\renewcommand{\arraystretch}{1.5}
\setlength\tabcolsep{8pt}
\small

\scalebox{0.95}{
\begin{tabular}{ lc } 
\hline
\textbf{Framework} & \textbf{Average Inference Time}\\ \hline
SeperateRAG (bge-zh-large) & 0.65s  \\
BSharedRAG & 0.91s  \\ 
\bottomrule
\end{tabular}
}
\caption{
We calculate the average inference time over 1000 questions randomly sampled from WorthyBuying Dataset. Both BSharedRAG retriever and generator are accelerated by vllm \cite{kwon2023vllm}
}.
\vspace{-2mm}
\label{table:time}
\end{table}

%% file: tex/07-conclusion_and_appendix.tex
\section{Conclusion}
In this paper, we construct a high-quality e-commerce dataset called WorthBuying, which contains an e-commerce knowledge base and QA pairs for training and evaluating retrieval and generation effectiveness.
We also design BSharedRAG, which can effectively adapt RAG models to a specific domain. In our framework, we propose sharing the backbone model that benefits from domain-specific continual pre-training and using hard negative mining and RAG instruction tuning to optimize two Low-rank Adaptation (LoRA) modules in the retriever and generator respectively. Compared to the fully shared RAG framework, our method can avoid the issue of negative transfer between tasks and the difficulty of loss balancing. Experimental results demonstrate that compared to existing separate RAG frameworks, both the retrieval performance and the generation performance have been significantly improved. 
\newpage \newpage
\section* {Acknowledgements}
This work is supported by the National Key R\&D Program of China (2023YFF0905402) and ZHI-TECH GROUP. We would like to thank \href{https://www.zhidemai.com/}{Beijing Zhidemai Technology Co., Ltd.} for supporting their crucial technical and administrative support, along with the resources that greatly contributed to model development.

\section*{Limitations}
The limitations of this work are as follows: (1) Due to the limited budget, we currently use only Chinese models and datasets in e-commerce domain in our experiments. As the proposed method is language and domain independent, it shall be applicable to more. Thus we plan to extend our methods to more languages and domains. (2) We implement a fully shared RAG framework based on one LoRA in this work. We are interested to explore more possibilities to compare two frameworks.   
(3) Our approach focuses on a basic retrieval and generation architecture. In the future, we plan to explore more advanced methodologies, such as after-retrieval Chain-of-Thought, re-ranking, and query rewriting, to enhance RAG's capability and robustness in the e-commerce domain.
\section*{Ethical Statement}
Our work aims to adapt a general LLM to the E-commerce domain, but the models we train may have negative impacts. For example, they could be used inappropriately, although we have performed data cleaning to avoid offensive content. However,
this is a common issue currently faced in the LLM field, and it is not amplified by this work. In the future, we will consider more work on the safety of LLMs to optimize their security in the E-commerce domain.
To protect the intellectual property rights of the data, we will strictly limit the dissemination of this dataset to academic research purposes only.


\bibliography{anthology, custom,ecomrag, pqa}

\newpage
\onecolumn
\appendix
\label{sec:appendix}
\section{Details of WorthBuying Dataset}
\subsection{Examples}
We show two examples of our constructed WorthBuying Dataset.
\begin{tcolorbox}
\textbf{title}: Upgraded from Xiaomi's first robot vacuum to the Roborock G10S—here's my honest take.  \\

\textbf{contents}: \\
In today's digital landscape, authentic product reviews are a rarity amidst the overwhelming amount of marketing content. Motivated by this, I've decided to share my genuine experience with the original Xiaomi and Roborock G10s vacuum robots to help potential buyers get a clearer picture.\par
~~~~Roborock is praised online for its advanced algorithms and reliable quality. Conversely, Ecovacs is often criticized for poor algorithms and quality, despite being the top offline seller. Xiaomi and other similar brands are noted for their cost-effectiveness but lack distinctive features. I purchased the first-generation Xiaomi robot in 2017 for ¥1699, and it's still running well without any major part replacements except for the filter. The Roborock G10s, bought during a sale in 2023 for ¥3999 without any freebies, was intended for my 101m² apartment with multiple rooms and balconies.\par
Mapping issues occurred with the Roborock during the first two attempts, taking over 17 minutes each, but it succeeded on the third try in 11 minutes. Both robots struggle with certain thresholds and may require virtual barriers if doors are not closed. Cleaning efficiency seems stagnant over the years. The Roborock did not significantly outperform Xiaomi despite newer technology, spending 65 minutes to clean a slightly smaller area than Xiaomi's 60 minutes for a similar space. Obstacle avoidance is one area where Roborock excels, identifying and navigating around small objects like cables and tissues, which Xiaomi tends to run over. \par
The mopping function of the Roborock was adequate, though it left water streaks, and the overall cleaning process was lengthy, especially when using separate sweeping and mopping modes. The Roborock app is user-friendly, offering customizable cleaning modes and map management, although it occasionally fails to retrieve maps.\par
~~~~In summary, while Roborock recognizes more obstacles and has a better mopping system, the difference in everyday use is minimal. Both require preliminary floor clearing, and neither has a significant edge in cleaning area or speed. For those with limited budgets prioritizing cleaning effectiveness, other models like the T7s Plus might be a viable, more affordable alternative.\\

\textbf{type\_one}: Home appliances \\

\textbf{type\_two}: Home devices \\

\textbf{cnt\_details}: \\
\{ quality\_score:2, share\_count: 3, like\_count: 8,, collect\_count: 10, comment\_count: 9 \} \\

\textbf{user\_question}: What are the differences between my original Xiaomi robot vacuum and the Roborock G10S in terms of usage? \\

\textbf{answer}: The Xiaomi robot primarily focuses on sweeping and has less advanced obstacle detection, often running over small objects. In contrast, the Roborock G10S offers better obstacle avoidance, customizable cleaning through its app, and combines sweeping with more effective mopping, though it initially struggles with accurate mapping and leaves more water residue.
\end{tcolorbox}
\newpage
\begin{tcolorbox}
\textbf{title}: Upgrading my new PS5 with a 2TB Fantom S790 SSD due to the price drop!  \\

\textbf{contents}: \\
I recently got a PS5 Digital Edition and quickly realized the storage was a bit tight after installing a few games. I took the opportunity to sell my idle PS4 to get some cash back. Meanwhile, with SSD prices plummeting, I decided to go all out and bought the 2TB Fantom S790 solid-state drive. \\
The drive comes in traditional Chinese black packaging with a highly recognizable Peking opera mask design. The back of the package lists the specific parameters and product information, offering a five-year warranty. Inside the box, there's the SSD itself, a small screwdriver, a screw, and a user manual.
The drive has a graphene cooling sticker. If the heat sink uses a thermal silicone sheet, you'll need to remove the sheet before installation. The back of the drive also has a sticker with basic parameter info for easy reference. \\
Installation steps:
\begin{enumerate}
    \item Before opening your console, lay something underneath to avoid scratches, then gently open the side of the case without the logo.
    \item Use a Phillips screwdriver to open the SSD compartment at the bottom, which shows different mounting positions depending on the motherboard.
    \item The screw and mount are fixed at the 110 position. Loosen the screw there.
    \item This SSD is of the 2280 standard size, so set the mount at the 80 position.
    \item Install the SSD, tighten the screw, reset the console to test it, then put the case back on.
\end{enumerate}
Testing:
\begin{enumerate}
    \item Power on, and the system directly detects the SSD and prompts for formatting.
    \item I'm not sure if the speed is an issue.
    \item After formatting, go to settings to select the default installation location for games.
    \item You can choose the default installation location.
    \item It shows a total of 2TB capacity, which is accurate\\
\end{enumerate}

\textbf{type\_one}: Computer and Technology \\

\textbf{type\_two}:Computer Accessories \\

\textbf{cnt\_details}: \\
\{ quality\_score:3, share\_count: 4, like\_count: 5,, collect\_count: 27, comment\_count: 15 \} \\

\textbf{user\_question}: Can I expand the storage on my Sony PlayStation 5 with a Fantom S790 SSD? \\

\textbf{answer}: Yes, you can use the Fanxiang S790 solid-state drive to expand the storage space of the Sony PlayStation 5 console. It should be noted that you need to follow the official tutorial and make sure that the solid-state drive used meets the requirements and standard specifications of the PS5.
\end{tcolorbox}
\newpage
\subsection{QA Generation with the help of GPT-4 API}
In this study, we employ a methodology based on an assessment of a product review article to simulate user queries from the perspective of GPT-4. Utilizing the in-context learning approach~\cite{ram2023incontext}, GPT-4 generates questions that mimic potential user concerns and queries. Subsequently, we apply a similar technique to enable the model to respond to these generated questions using information derived from the original article.

\subsubsection{Question Generation Prompt} 

\begin{tcolorbox}
Assume you are a customer purchasing goods on an e-commerce platform, and you have a question about a product. After reading the article, your question is well answered. What is your question?\\

Requirements: The question should be concise and end with a question mark "?"\\

[Example 1]

\textbf{Article}:

Practical test of the Xiaomi 65w gallium nitride charger charging a Samsung Note10+: Update modification (2020-03-06 17:53:12): Conclusion, the Xiaomi 65w gallium nitride does not support the 45w charging of Samsung Note10+. Among the several PPS-supporting PD chargers tested, the highest can reach the same 25w as the original charger.Xiaomi 65w Gallium NitrideWithout further ado, let’s directly look at the pictures for yourself; the phone was in a screen-off state with about 35\% battery level, using the original charger cable. For a few that did not have the original cable, I used the Lenovo lipstick cable.Baseus 65w Gallium Nitride \\
Green Union 30w PD \\
Anker 18w nano PD \\
Lenovo lipstick 65w PD \\
Set as main image \\
Xiaomi 65w PD (actually the same as the ZMI black 65w) \\
Samsung Note10+ original 25wPD

\textbf{Your question is}:

Can I use the Xiaomi 65w gallium nitride charger to charge my Samsung Note10+?\\

[Example 2]

\textbf{Article}:

\{passage\}

Your question is:

\end{tcolorbox}

\newpage
\subsubsection{Answer Generation Prompt} 
\begin{tcolorbox}
As a question-and-answer AI specializing in the e-commerce sector, I am equipped to answer user inquiries strictly based on the reference material provided. There are no scenarios in which I am unable to provide an answer, as all questions directly originate from the supplied references. Please rely on the information provided for accurate responses.\\

[Example 1]

\textbf{Article}:

Practical test of the Xiaomi 65w gallium nitride charger charging a Samsung Note10+: Update modification (2020-03-06 17:53:12): Conclusion, the Xiaomi 65w gallium nitride does not support the 45w charging of Samsung Note10+. Among the several PPS-supporting PD chargers tested, the highest can reach the same 25w as the original charger.Xiaomi 65w Gallium NitrideWithout further ado, let’s directly look at the pictures for yourself; the phone was in a screen-off state with about 35\% battery level, using the original charger cable. For a few that did not have the original cable, I used the Lenovo lipstick cable.
Baseus 65w Gallium Nitride \\
Green Union 30w PD \\
Anker 18w nano PD \\
Lenovo lipstick 65w PD \\
Set as main image \\
Xiaomi 65w PD (actually the same as the ZMI black 65w)Samsung Note10+ original 25wPD.

\textbf{question }:

Can I use the Xiaomi 65w gallium nitride charger to charge my Samsung Note10+?

\textbf{Answer:}

You can use the Xiaomi 65w Gallium Nitride charger to charge your Samsung Note10+ smartphone; however, it is important to note that this charger does not support the 45w fast charging capability of the Samsung Note10+. According to test results, when utilizing the Xiaomi 65w Gallium Nitride charger, the maximum charging power attained is 25w, which is equivalent to the charging performance of the original Samsung Note10+ 25w PD charger. Therefore, although it cannot achieve the maximum charging speed of 45w, it can still safely charge at a rate of 25w.\\

[Example 2]

\textbf{Article}:

\{passage\}

\textbf{question}:

\{question\}

\textbf{Answer:}
\end{tcolorbox}
\newpage
\begin{figure}
    \centering
    \includegraphics[width=0.9\linewidth]{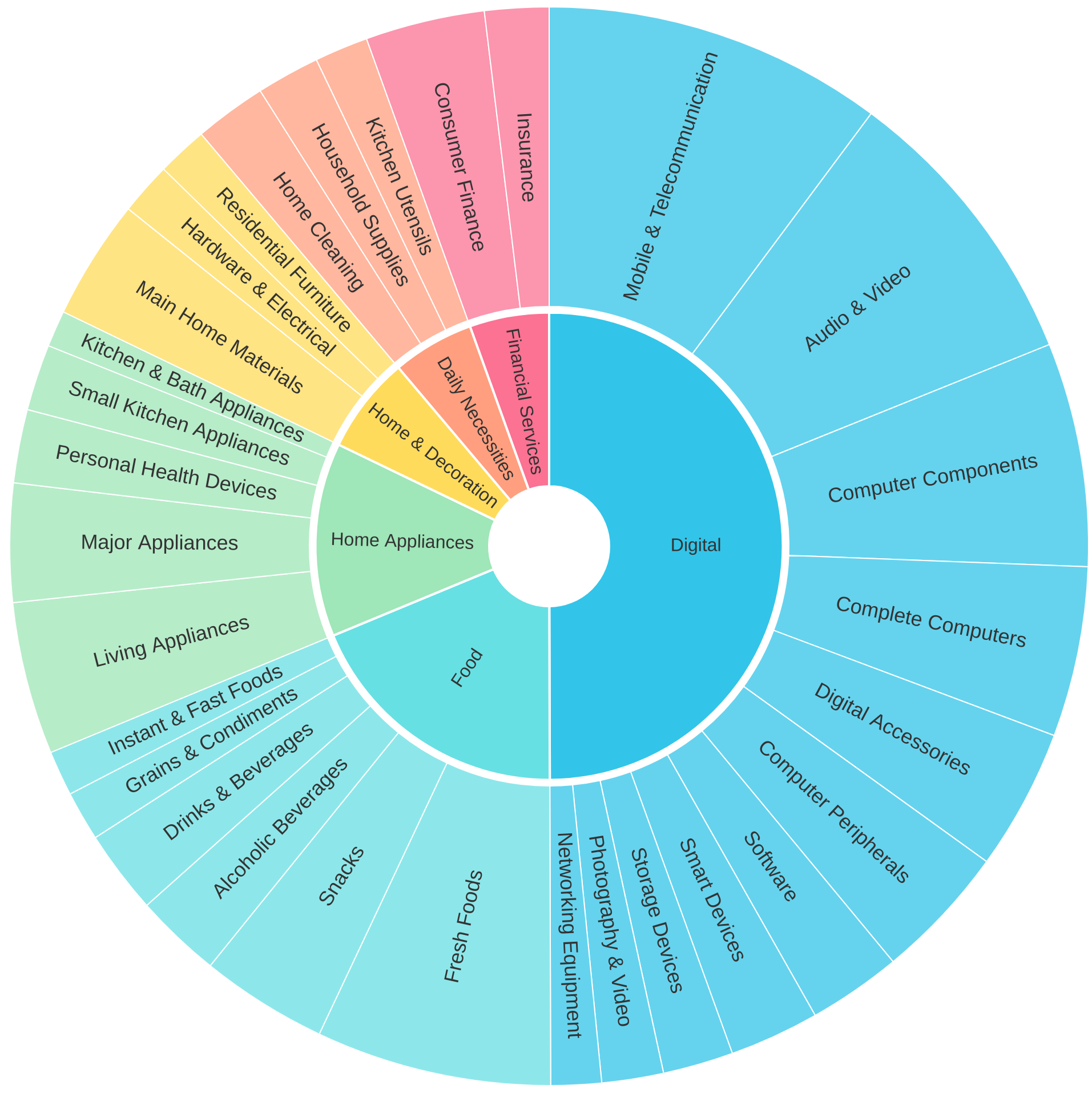}
    \caption{Partial categories of WorthBuying dataset}
    \label{fig:enter-label}
\end{figure}

\section{Model Training Details}
\label{sec:appendix_data_details}
\subsection{Domain-specific Continual Pretraining}
In this section, we describe the continual pretraining process for the Baichuan2-7B-Base model \cite{yang2023baichuan} and  the recipe of training data.  Our models are trained on Llama-Factory \cite{zheng2024llamafactory} framework.
\input{tables/pretrain_data}
\subsubsection{Training Dataset}
The detailed training datasets are listed in Table \ref{table:pretrain}. For the e-commerce domain dataset, besides our WorthBuying dataset we collect e-commerce relevant notes from Xiaohongshu and Zhihu websites. For the generation domain data, we use the open-source pertaining dataset from Linly-OpenLLaMA \footnote{\href{https://github.com/CVI-SZU/Linly}{https://github.com/CVI-SZU/Linly}} and WanJuan Corpus \cite{qiu2024wanjuancc}

\subsubsection{Training Configurations}
The model is trained using the DeepSpeed framework, which facilitates efficient distributed training. Key hyperparameters and configurations include:
\begin{itemize}
    \item \textbf{Precision:} Training was performed using bfloat16 mixed precision, which significantly reduces memory consumption while maintaining the model's performance.
    \item \textbf{Optimizer:} We used the AdamW optimizer with a dynamically adjusted learning rate based on the training progression. This setup helps in stabilizing the training process in the initial phases and gradually fine-tuning the weights in later stages.
    \item \textbf{Learning Rate Scheduler:} A cosine decay scheduler with a warmup phase was implemented. The learning rate started at $2 \times 10^{-5}$ and was adjusted according to a predefined schedule to optimize convergence.
    \item \textbf{Model Parallelism:} We employed Zero Stage 2 optimization to manage GPU memory efficiently, offloading optimizer states and model parameters to CPU memory. 
    \item \textbf{Hardware Configureation} The training was conducted on a cluster of four NVIDIA H100 80GB GPUs,
    \item \textbf{Other Configurations}   We set 32 per device batch-size and use 100 step for warming up.
    
\end{itemize}

\subsection{Contrastive Learning Training Details}
In our experiments, we utilize DeepSpeed for distributed training with four Nvidia-H100 80GB GPUs. Key hyperparameters included mixed precision training with bf16, a batch size of 4 per device with 4 gradient accumulation steps, and gradient checkpointing.  The model is trained for one epoch with a learning rate of 1e-5, logging every 5 steps, and overwriting the output directory. Dataset processing uses 32 parallel processes, with cross-device negative sampling and a 100-step warmup phase to stabilize training. 

In the training dataset, we integrate some open-source retrieval datasets with our WorthBuying dataset. For the open-source dataset, we use the train dataset of the retrieval part of C-MTEB \cite{bge} Benchmark.

\section{Generation Evaluation Details}
In the evaluation of generation, as mentioned in \cite{es2023ragas}, due to the subjective nature of responses generated by models, it is challenging to directly compute accuracy. Therefore, we adopt the multi-turn evaluation method utilized in \textit{ragas} leveraging GPT-4. Below, we outline the prompts used in our evaluation to provide clarity and transparency in our assessment process.
\label{sec:expr_details}
\subsection{ Stage1: Extracting key points from answer}
\begin{tcolorbox}
Below are questions and answers, from which key points have been extracted and presented as individual statements.\\
======================\\

Question: What are the specifications of the Huawei X 14 smartphone?\\

Answer: Storage options include 256G and 512G; screen resolution is 2640x1080 FHD+ with a peak brightness of 1750 nits and a refresh rate of 120Hz; the internal screen measures 6.7 inches with a square aspect and 6.6 inches with rounded corners, while the external screen measures 3.4 inches square; it utilizes the Snapdragon 8 Gen 2 mobile platform with an octa-core CPU; the weight is a steady 187g; battery capacity is 3700mAh.\\

Statements:

Storage options of 256G and 512G.\\
Screen resolution of 2640x1080, refresh rate of 120Hz\\
Internal screen size of 6.7 inches, external screen of 3.4 inches square\\
Uses Snapdragon 8 Gen 2 mobile platform, CPU is octa-core\\
Weighs 187g.\\
Battery capacity is 3700mAh.\\
======================\\

Question: What oral supplements can help women restore their skin?\\

Answer: Vitamin C tablets from Northeast Pharmaceuticals, Kang'enbei's Vitamin C chewable tablets, and Northerland's compound Vitamin B and Vitamin C chewable tablets can help restore women's skin. Vitamin C can whiten the skin and regulate the body, while Vitamin B can repair the skin and is effective in treating acne and mouth ulcers.\\

Statements:\\

Vitamin C can whiten and help restore women's skin.\\
Vitamin B can repair the skin and is effective for acne and mouth ulcers.\\

======================\\

Question: Can the refill in a Baixue ballpoint pen be replaced?\\
Answer: Sorry, I cannot answer this question based on the available information.\\

Statement:\\

The model is unable to answer this question.\\

======================\\

Question: \{question\}\\
Answer: \{answer\}\\
Statement:\\
\end{tcolorbox}

\subsection{ Stage2: Judging the key points}
\begin{tcolorbox}
Below are a series of statements along with corresponding facts. Each statement is evaluated for honesty, determining if it is sufficiently supported by the facts provided.\\
======================\\
Statement: \\
1. Storage capacity of 256 and 512GB.\\
2. Uses the Snapdragon 8 Gen 2 mobile platform, CPU has nine cores.\\
3. Weighs 199g.\\
4. Battery capacity is 3700mAh.\\

Fact: \\
This smartphone offers storage capacities of 256 and 512GB with an upgrade option available. Screen resolution is 2640x1080 FHD+ with a peak brightness of 1750 nits and a refresh rate of 120Hz; the internal screen sizes are 6.7 inches square and 6.6 inches rounded, external screen size is 3.4 inches square; it is equipped with the Snapdragon 8 Gen 2 mobile platform, and the CPU has eight cores; the weight is stable at 199g; battery capacity is 4600mAh.\\

Analysis:\\
1. Storage capacity of 256 and 512GB. This statement matches the facts, correct. [Yes]\\
2. Uses the Snapdragon 8 Gen 2 mobile platform, CPU has nine cores. According to the facts, the CPU should have eight cores, this statement is factually incorrect. [No]\\
3. Weighs 199g. This is consistent with the facts, correct. [Yes]\\
4. Battery capacity is 3700mAh. This does not match the facts, which state the battery capacity should be 4600mAh, incorrect. [No]\\

=====================\\

Statement: \\
1. The model does not have enough information and cannot answer the question, the model's response is unknown.\\

Fact: \\
Vitamin C tablets from Northeast Pharmaceuticals, Kang'enbei’s Vitamin C chewable tablets, and Northerland's compound Vitamin B and Vitamin C chewable tablets can help restore women's skin. Vitamin C has a whitening effect and can regulate the body, while Vitamin B can repair the skin and is effective in treating acne and mouth ulcers.\\

**Analysis:** \\
Although the model states it does not know, this makes it impossible for me to judge the response based on the provided facts. [Unknown]\\

=====================\\

Statement: \\
\{statement\}\\

Fact: \\
\{ground\_truth\}\\

Analysis:
\end{tcolorbox}

\section{Additional Analysis}

\begin{figure}[t]
    \centering
    \includegraphics[width=0.85\linewidth]{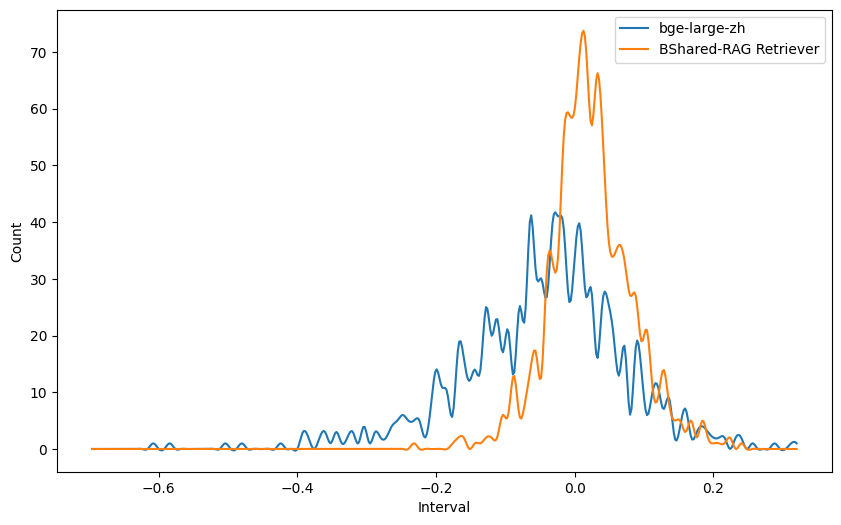}
    \caption{Histograms of gap scores, calculated by the similarity score of ground-truth document minus the similarity score of top returned not-ground-truth document, for two retrievers. The more positive gap scores the better.}
    \label{fig:gap}
\end{figure}

BShareRAG retriever can better distinguish hard negatives. We analyze the different behavior between two retrievers, \ie, BGE-large-zh and BShared-RAG Retriever. For each question, we first calculate the gap of similarity scores between a ground-truth document (it surely contains answers) and the top returned not-ground-truth document. The gap score ranges from -1 to 1. A larger gap means a retriever can better rank the ground-truth document among others. Then we draw histograms of the gap scores for the two retrievers, as shown in Figure~\ref{fig:gap}. It indicates that our BShared-RAG retriever can better distinguish the ground-truth document from other similar documents as the gap scores are positive for more questions. In contrast, \textit{BGE-large-zh} has a more flat distribution, which means it cannot estimate a larger similarity of the ground-truth document for many questions.


%% file: tables/pretrain_data.tex
\begin{table*}[ht]
\centering

\setlength\tabcolsep{3.5pt}

\begin{tabular}{ccccc}

\toprule

DataType & Language/Source & Weight & Number of Files & Size(GB) \\
\midrule
\multirow{3}*{E-commerce Domain dataset} & WorthyBuying Dataset & 2.53\% & 735,975 & 2.43GB\\
~ & Xhs-ecommerce-note & 4.96\% & 2,841,686 & 4.5GB\\
~ & Zhihu-ecommerce-note & 7.91\% &2,996,802 & 7.59GB\\
\midrule
\multirow{6}*{General dataset} & ZH-Wiki & 2.60\% & 1,348,766 & 2.49GB\\
~ & Common-crawl-web & 20.29\% & 6,741,652 & 19.44GB\\
~ & Chinese-new & 16.14\% & 5,208,071 & 15.46GB\\
~ & Chinese-Law & 28.21\%& 5,363,805& 27.02GB\\
~ & Chinese-Patents& 15.79\% & 1,127,932& 15.13GB\\
~ & Chinese-CLS & 1.78\%& 2,310,165& 1.71GB\\
\bottomrule
\end{tabular}

\caption{Training data of Continual pretraining}
\label{table:pretrain}
\end{table*}